# An Once-for-All Budgeted Pruning Framework for ConvNets Considering Resolution

Wenyu Sun, Jian Cao, Pengtao Xu, Xiangcheng Liu, Pu Li
Peking University
Beijing
sunwenyu@pku.edu.cn

## Abstract

*We propose an efficient once-for-all budgeted pruning framework (OFARPruning) to find many compact network structures close to winner tickets in the early training stage considering the effect of input resolution during the pruning process. In structure searching stage, we utilize cosine similarity to measure the similarity of the pruning mask to get high-quality network structures with low energy and time consumption. After structure searching stage, our proposed method randomly sample the compact structures with different pruning rates and input resolution to achieve joint optimization. Ultimately, we can obtain a cohort of compact networks adaptive to various resolution to meet dynamic FLOPs constraints on different edge devices with only once training. The experiments based on image classification and object detection show that OFARPruning has a higher accuracy than the once-for-all compression methods such as US-Net and MutualNet (1-2% better with less FLOPs), and achieve the same even higher accuracy as the conventional pruning methods (72.6% vs. 70.5% on MobileNetv2 under 170 MFLOPs) with much higher efficiency.*

## 1. Introduction

Convolutional neural networks have been widely used in image classification, object detection, semantic segmentation and many other fields. In addition to model accuracy, the security of data during transmission has also attracted increasing attention. Therefore, more models should be deployed to edge devices in practical application. However, with the improvement of network accuracy, the size of the network becomes larger, requiring high amounts of storage space and computing resources, making it difficult to deploy the network to edge hardware with data security but resource constraints. While ensuring the accuracy of the model and reducing the amount of model parameters and computational consumption as much as possible, researchers have proposed several model compression methods, such as model pruning [18, 21, 22, 28, 9], quantization [38, 15, 25, 5], network architecture search [20, 7, 36, 29, 31, 17, 14] and lightweight network design [26, 24, 8, 16]. Pruning has been proven to be an effective way to reduce model parameters for accelerating inference which can be divided into unstructured pruning and structured pruning. The former lacks versatility due to the requirement of a corresponding sparse matrix calculation acceleration library when deploying to the edge devices. The latter directly prunes the kernels, which can achieve acceleration on common hardware.

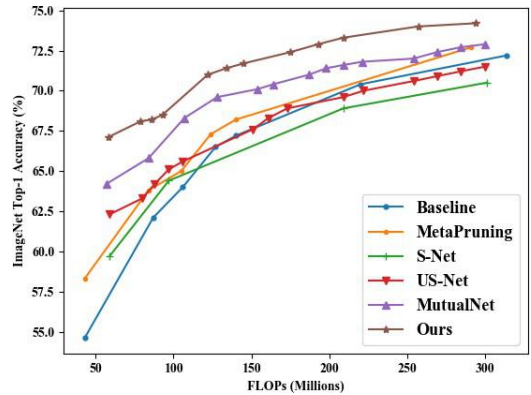

Figure 1. Overall comparison of various compression methods based on MobileNetv2. Among them, MetaPruning is once training for one network, the rest is once-for-all compression like ours.

Since the lottery hypothesis [6] was proposed, iterative pruning to find the optimal compact network has become the mainstream method of structured pruning. It requires time-consuming iterative training, pruning and fine-tuning to get a network that satisfies the FLOPs constraints of edge devices. And it is often necessary to achieve accuracy-latency tradeoff in actual applications resulting in a quite long development cycle. Besides, the large feature map of the first few layers of the network occupies a high proportion of FLOPs, and for some special hardware such as spiking neural network chips[1], a large feature map will cause network deployment difficulties. In addition, we find that knowledge distillation is necessary for compact network with a large pruning rate to improve accuracy. Based on the above findings, we propose a simple and efficient budgeted pruning framework, which randomly sample networks with different pruning rates to form a teacher-teaching assistant-student structure based on sandwich rule and utilize teaching assistant inplace distillation to search the networks corresponding to different pruning rates close to winner tickets in above framework. After searching, we adopt different resolution input for the aforementioned structures to perform joint optimization, so as to achieve multiple networks that are adaptive to different resolution to meet the dynamic constraints of FLOPs. In summary, our main contributions are as follows:

1. Propose a simple and effective neural network pruning framework which considers the effect of input resolution during the pruning process, and solve the problem that large feature maps hold numerous computation to make it feasible to deploy such networks to the specific hardware.

2. In above framework, we use an efficient method to search the networks with different pruning rates close to winner tickets. After searching, the aforementioned networks are jointly optimized through random sampling and teaching assistant inplace distillation to obtain a cohort of compact networks to meet the dynamic constraints of FLOPs which greatly reduces the time and energy consumed. From the perspective of pruning, our framework achieves once training for many compact networks.

3. Analyze the optimal configuration of the above pruning framework through ablation experiments, and prove the effectiveness of the pruning framework through image classification and object detection experiments. It shows that our framework achieves a higher accuracy than the once-for-all compression methods and achieves the same even higher accuracy as the conventional pruning methods with higher efficiency.

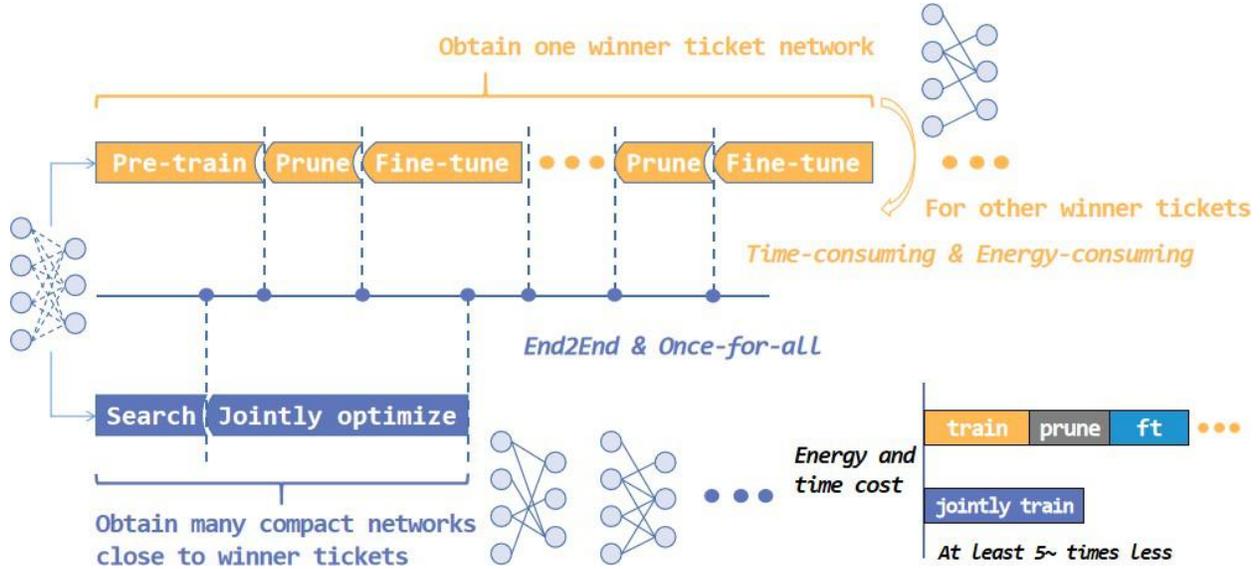

Figure 2. Comparison between conventional pruning methods based on lottery hypothesis and our proposed method. Conventional methods need pruning, fine-tuning and iterate this process to get a compact network while our method can obtain multiple compact networks close to winner tickets after once search and once joint training, saving a lot of energy and time.

## 2. Related work

Due to acceleration on common hardware platforms, structured compression methods have attracted more and more attention. The mainstream structured compression methods can be divided into changing the network width through scaling factors and structured pruning. The former has higher development efficiency and simplicity, but the compact model is not necessarily optimal. The latter ranks the importance of the convolutional kernels in the pretrained network according to a certain standard, and then prunes the unimportant kernels so as to reduce the parameters and speed up the network inference. The conventional pruning methods can only obtain single compact network through once training while recent work focuses on how to obtain many compact models through once training to minimize time and energy consumption.

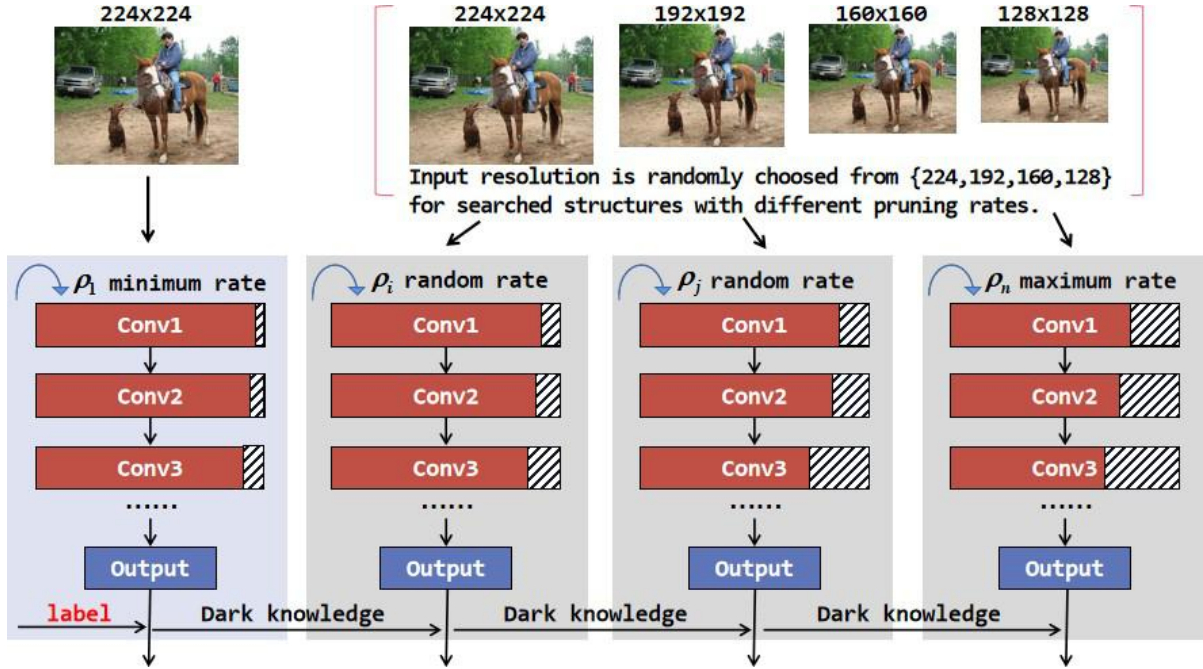

Figure 3. The frame of our proposed OFARPruning. The pruning rate is sampled from $\rho_1, \rho_2, ..., \rho_n$ to meet the dynamic resource constraints of different edge devices and input resolution is sampled from {224, 192, 160, 128} in each iteration. We follow the sandwich rule to sample four networks with different pruning rates. For the first network, we adopt minimum pruning rate and feed 224 resolution to it due to its strong representation ability. For the other networks, we randomly sample the resolution for them. The first network is optimized by the ground truth, the other networks are optimized by the output of previous network to eliminate the representation gap between networks with different pruning rates. When the pruning mask similarity of a compact structure exceeds the threshold, the structure will be stored in the pruned network pool (PNP) for the following training.

**Once training for one network.** NetworkSlimming [21] sparse the gamma of the BN layers by introducing a sparse factor in training process and then prune the kernels corresponding to the BN layers with small gamma values. ThiNet [23] minimize the reconstruction error to determine which channel to prune in the current layer through the statistics of the next layer based on the greedy strategy. FPGM [12] regards the convolutional kernels as points in Euclidean space and calculate the geometric median of each convolution layer to prune the kernels close to the geometric median for model compression. HRank [19] achieves pruning by utilizing the average rank of the output feature map. MetaPruning [22] generates weights for encoding vectors of different structures by training a PruningNet to find a optimal network structure under the certain constraint of FLOPs.

**Once training for many networks.** S-Net [37] and US-Net [35] achieve joint training of networks with different widths to obtain multiple sub-networks at one time through weight sharing between networks and outstanding statistical methods for BN layers. MutualNet [30] considers the input resolution as an influencing factor to reduce FLOPs on the basis of the network width, and achieve the joint training of network width and input resolution. OFA [2] pretrains a network then obtains sub-networks through progressive shrinking and fine-tuning to achieve once training for all networks.

## 3. Methodology

### 3.1. Preliminary

**Sandwich Rule for OFARPruning Frame.** In US-Net [35], the joint optimization of networks with arbitrary width is achieved by accumulating loss from sub-networks during training. In our proposed framework, we sample the minimum pruning rate network, the maximum pruning rate network and two networks with random pruning rates in each iteration. As shown in Equation 1, $n$ means the channel number with the minimum pruning rate and $k$ means the first $k$ channels of $n$ with other pruning rates, all randomly sampled pruned sub-networks can be affected by optimizing the minimum pruning ratio network (upper bound network) and the maximum pruning ratio network (lower bound net-

## Algorithm 1 OFARPruning

**Input:** Full model $f(x, W_0)$, Various edge deivces FLOPs constraints after sorting in descending order $(c_1, c_2, c_3, c_n)$, similarity threshold $\tau$
**Output:** many compact networks adaptive to different resolution

1: Calculate the pruning rate that satisfies the FLOPs constraints of different edge devices at different resolution $\{\rho_1, \rho_2, ,\rho_n\}$
2: Build the pruning frame as shown in Figure 3
3: **for** $epoch = 0, 1, \ldots, N_{search}$ **do**
4:     Prune full model with different pruning rate based on Network Slimming [21]
5:     Calculate mask similarity between networks in adjacent epoch.
6:     **if** mask similarity $> \tau$ **then**
7:         Add the corresponding structure to PNP
8: **End searching**
9: **for** iteration in every epoch **do**
10: Fix student and teacher network, the rest are randomly sampled from PNP. Choose maximum resolution input for teacher network and randomly choose resolution for other networks.
11: **End training**

work).

$$(1)_{0 \le \varepsilon \le \varepsilon^n} \underset{k}{\le} \varepsilon, \varepsilon = \frac{v}{e}$$

**Inplace Distillation with Teaching Assistant.** Knowledge distillation is regarded as an effective method for improving the accuracy of compact network by transferring teacher network information to the student network which is widely used in practice. However, it is found that when the teacher network and the student network have a large gap in information extraction ability, the student network cannot effectively learn the large amount of detailed knowledge of the teacher network. Therefore, a medium size network is used as a teaching assistant network between the teacher network and the student network to process the information. Therefore, in our proposed framework, we form a teacher-teaching assistant-student structure to make up for the gap. According to the ablation experiment, such distillation structure is better than ordinary distillation methods.

**Batch Normalization Statics.** Since OFARPruning can customize the number of compact networks, we can use the BN storage method in S-Net or the post-statistics method of BN in US-Net. When the edge devices resources do not change much dynamically, in other words, once training does not require many compact networks, the former statistical method can be used. The latter is more suitable when the resource range of the edge devices is large and many compact networks are required for once training.

Table 1. ImageNet performance comparison of different conventional pruning methods based on MobileNetv2 under same FLOPs constrain. All the results of our method are achieved through once training according to Fig. 3, in other words, OFARPruning is once-for-all.

| Network | FLOPs | Top-1 Accuracy |
|---|---|---|
| Full Model [26] | 314M | 72.2% |
| GFS [32] | 258M | 71.9% |
| GOPWL [33] | 245M | 72.2% |
| **OFARPruning(Ours)** | 253M | **74.0%** |
| Uniform [26] | 220M | 69.8% |
| AMC [11] | 220M | 70.8% |
| LeGR [4] | 224M | 71.4% |
| GFS [32] | 220M | 71.6% |
| MetaPruning [22] | 217M | 71.2% |
| GOPWL [33] | 218M | 71.7% |
| **OFARPruning(Ours)** | **209M** | **73.3%** |
| ThiNet [23] | 175M | 68.6% |
| DPL [39] | 175M | 68.9% |
| GFS [32] | 170M | 70.4% |
| GOPWL [33] | 170M | 70.5% |
| **OFARPruning(Ours)** | 179M | **72.6%** |
| Uniform [26] | 106M | 64.0% |
| MetaPruning [22] | 105M | 65.0% |
| GFS [32] | 107M | 66.9% |
| **OFARPruning(Ours)** | 107M | **69.2%** |

### 3.2. Rethink Network Budgeted Pruning

Budgeted pruning is defined as pruning networks under resource constraints. Since the lottery hypothesis was proposed, iterative pruning has become the mainstream method used in budgeted pruning to ensure high compression rate and smalanzl accuracy loss. As shown in Fig. 2, the lottery hy- pothesis based pruning method requires pretraining, prun- ing, fine-tuning and then iterate, which is time-consuming and energy-consuming. Early-Bird [34] found that the prun- ing mask is basically unchanged after a few epochs training and this phenomenon is common in each task and network which greatly reduces the time and energy cost of the prun- ing method based on the lottery hypothesis. However, such method still requires fine-tuning for each searched network structure. When there are multiple edge devices constrained by different FLOPs, it will still consume a lot of time and energy.

Besides, most of the pruning methods only focus on the structure of the network, and neglect the effect of input resolution on FLOPs. As shown in Equation 2, where $K$ represents kernel size, $W$ and $H$ represent width and height of output feature map, $C_{in}$ and $C_{out}$ represent input channel and output channel number, when the structure is close to the collapsed node, in other words, keeping pruning the structure will cause a lot of accuracy drop, pruning the convolutional kernels is not the optimal choice. Instead, the input resolution can be changed to reduce FLOPs and maintain ordinary accuracy.

Table 2. ImageNet performance comparison of different conventional pruning method based on ResNet50 under same FLOPs constrain. All the results of our method are achieved through once training, in other words, OFARPruning is once-for-all.

| Network | FLOPs | Top-1 Accuracy |
|---|---|---|
| ThiNet [23] | 2.9G | 75.8% |
| MetaPruning [22] | 3.0G | 76.2% |
| **OFARPruning(Ours)** | **2.8G** | **76.9%** |
| Uniform [10] | 2.3G | 74.8% |
| ThiNet [23] | 2.1G | 74.7% |
| CP [13] | 2.0G | 73.3% |
| FPGM [12] | 2.3G | 75.6% |
| HRank [19] | 2.3G | 75.0% |
| MetaPruning [22] | 2.0G | 75.4% |
| DCP [39] | 1.8G | 75.0% |
| **OFARPruning(Ours)** | **1.7G** | **76.1%** |
| Uniform [10] | 1.1G | 72.0% |
| ThiNet [23] | 1.2G | 72.1% |
| MetaPruning [22] | 1.0G | 73.4% |
| **OFARPruning(Ours)** | **1.0G** | **74.1%** |

$$FLOPs = 2K^2 \times W \times H \times C_{in} \times C_{out} \quad (2)$$

### 3.3. OFARPruning Frame

Aiming at the shortcomings of above iterative pruning methods, we propose a simple and effective neural network pruning framework shown in Fig. 3. The specific process of our proposed method is shown in Algorithm 1. In the search stage, our framework searches for the network structures close to winner tickets corresponding to different pruning rates. For the edge device $E_i(i = 1, 2, \cdots n)$, the network computation which meets the real-time requirements is $c_i$ MFLOPs, and the computation of the unpruned model is $C$ MFLOPs, then we can obtain the rough pruning rate of the network adaptive to the certain edge device is $\rho_i$. For each epoch in training process of the framework, we calculate the similarity of pruning mask between the current and last epoch according to Equation 3, where $a$ means the pruning mask in last epoch and $b$ means the pruning mask in current epoch. When the consequent similarity exceed the threshold, we add the structure to the Pruned Network Pool (PNP), and then the same method is used until all the compact networks are found. Figure 4 shows that in our proposed method, the pruning mask will effectively converge under different pruning rates after a few epochs of training.

$$similarity = \frac{\sum_{i,j=1}^{n} a_{ij} b_{ij}}{\sqrt{\sum_{i,j=1}^{n} a_{ij}^2} \cdot \sqrt{\sum_{i,j=1}^{n} b_{ij}^2}} \quad (3)$$

After the architecture search, the framework performs joint optimization on all structures by randomly sampling and makes each compact network consider the effect of input resolution during the training process. With the advantage that teacher network is able to extract detailed information of high-resolution images and the student network is more suitable for extracting general information of low-resolution images, we select 224 input resolution for the teacher network and randomly sample resolution from $\{224, 192, 160, 128\}$ for the rest networks. Among them, the teacher network is trained by ground truth, and the output of each subsequent network is used as the label of the next network which constitutes the teacher-teaching assistant-student architecture to solve the problem that the expression ability gap of searched networks corresponding to different pruning rates is too large. The loss calculation in the framework is as follows where $ttT$ means ground truth, $f(\cdot)$ means output of certain network, $\rho$ means pruning rate while $\rho_1$ means minimum pruning rate and $\rho_n$ means maximum pruning rate, $I_R$ means input resolution.

$$loss_T = -\frac{1}{N} \sum_{m=1}^{N} ttT^{(m)} \log[f(\rho = \rho_1, I_R = 224)]$$
$$+ \sum_{m=1}^{N} (1 - ttT^{(m)}) \log[1 - f(\rho = \rho_1, I_R = 224)] \quad (4)$$

$$loss_{TA1} = -\frac{1}{N} \sum_{m=1}^{N} \sum f(\rho = \rho_i, I_R = r_i) * \log_2 \frac{f(\rho=\rho_i, I_R=r_i)}{f(\rho=\rho_1, I_R=224)} \quad (5)$$

$$loss_{TA2} = -\frac{1}{N} \sum_{m=1}^{N} f(\rho = \rho_j, I_R = r_j) * \log_2 \frac{f(\rho=\rho_j, I_R=r_j)}{f(\rho=\rho_i, I_R=r_i)} \quad (6)$$

$$loss_S = -\frac{1}{N} \sum_{m=1}^{N} f(\rho = \rho_n, I_R = r_k) * \log_2 \frac{f(\rho=\rho_n, I_R=r_k)}{f(\rho=\rho_j, I_R=r_j)} \quad (7)$$

$$loss_{total} = loss_T + loss_{TA1} + loss_{TA2} + loss_S \quad (8)$$

Since we prune the networks based on global rank in our proposed framework, the method of calculating partial derivatives in MutualNet [30] can be reused to prove effectiveness of different resolutions. Assuming that the pruning rate of a sub-network is $\rho_1$ and the pruning rate of another sub-network is $\rho_2 (\rho_1 < \rho_2)$, the input resolution for the former is 128 while for the latter is 224. Since weights are shared between sub-networks, we can calculate the gradients as follows.

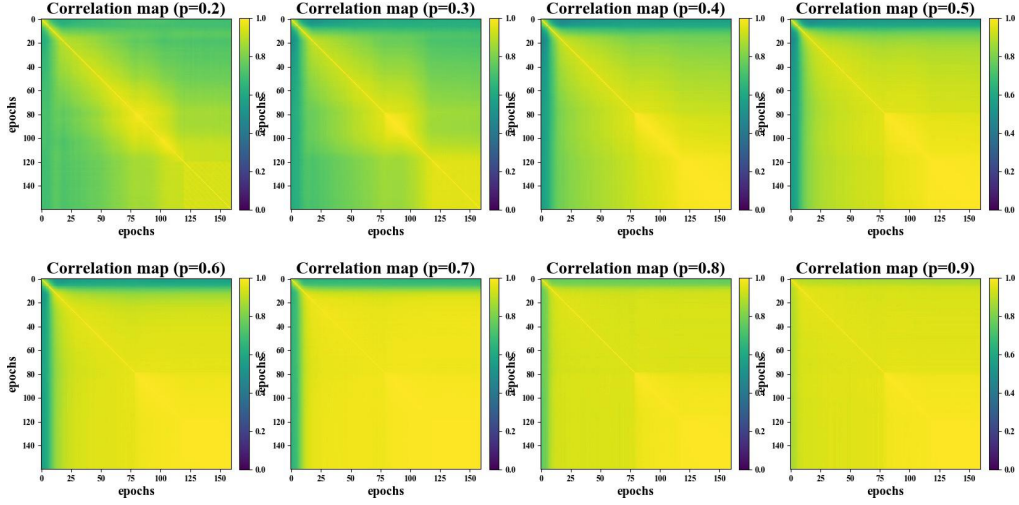

Figure 4. Similarity map with different pruning rates. The pruning mask will become very similar under a few epochs of training no matter what the pruning rate is.

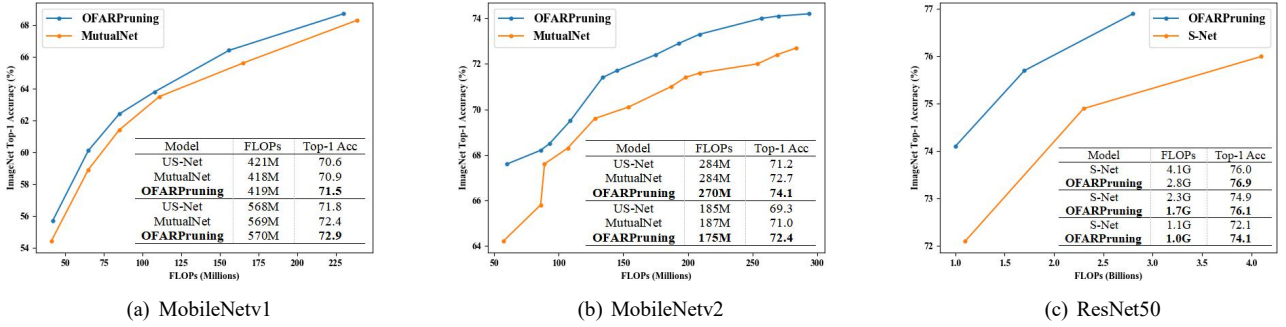

(a) MobileNetv1  (b) MobileNetv2  (c) ResNet50

Figure 5. ImageNet performance comparison of different once-for-all compression methods based on MobileNetv1, MobileNetv2 and ResNet50 under similar FLOPs constraint. The table in (a) shows some results not included in the curves.

$$\frac{\partial L_{W_{\rho_1}, I_R=224}}{\partial W_{\rho_1}} = \frac{\partial L_{W_{\rho_1}, I_R=224}}{\partial W_{\rho_2}} \oplus \frac{\partial L_{W_{\rho_1}, I_R=224}}{\partial W_{\rho_1-\rho_2}} \quad (9)$$

where $\oplus$ means vector concentration, $W_{\rho_1-\rho_2}$ means kernels which the former has while the latteisr doesn't. The total gradients are computed according to Equation 8 as follows.

$$\frac{\partial L}{\partial W} = \frac{\partial L_{W_{\rho_1}, I_R=224}}{\partial W_{\rho_1}} + \frac{\partial L_{W_{\rho_2}, I_R=128}}{\partial W_{\rho_2}}$$

$$= \frac{\partial L_{W_{\rho_1}, I_R=224}}{\partial W_{\rho_2}} \oplus \frac{\partial L_{W_{\rho_1}, I_R=224}}{\partial W_{\rho_1-\rho_2}} + \frac{\partial L_{W_{\rho_2}, I_R=128}}{\partial W_{\rho_2}}$$

$$= \frac{\partial L_{W_{\rho_1}, I_R=224} + \partial L_{W_{\rho_2}, I_R=128}}{\partial W_{\rho_2}} \oplus \frac{\partial L_{W_{\rho_1}, I_R=224}}{\partial W_{\rho_1-\rho_2}}$$

(10)

Since the first item in Equation 10 consist two parts including $I_R=224$ and $I_R=128$, it effectively proves that the sub-networks with different pruning rates have the ability to learn different resolution inputs in our proposed framework.

Table 3. COCO performance comparison of different once-for-all compression methods based on VGG-16 [27] backbone in MMDetection [3].

| Network | FLOPs | mean Average Precision |
|---|---|---|
| US-Net [35] | 110G | 34.4 |
| MutualNet [30] | 110G | 35.4 |
| **OFARPruning(Ours)** | **103G** | **36.1** |
| US-Net [35] | 45.9G | 33.0 |
| MutualNet [30] | 44.9G | 34.3 |
| **OFARPruning(Ours)** | **42G** | **35.0** |
| US-Net [35] | 2.5G | 25.0 |
| MutualNet [30] | 2.5G | 27.8 |
| **OFARPruning(Ours)** | **2G** | **29.1** |

Table 4. OFARPruning for single network Wider-ResNet based on CIFAR-10 and CIFAR-100. * means our implementation.

| Dataset | Network | FLOPS | Accuracy |
|---|---|---|---|
| CIFAR-10 | Baseline | 5243M | 96.1% |
| | NetworkSlimming* | 987M | 95.3% |
| | MetaPruning* | 903M | 95.6% |
| | **OFARPruning** | **828M** | **96.9%** |
| CIFAR-100 | Baseline | 5243M | 81.2% |
| | NetworkSlimming* | 1289M | 79.9% |
| | MetaPruning* | 1077M | 80.4% |
| | **OFARPruning** | **1022M** | **81.1%** |

## 4. Experiment

In this section, we prove the high efficiency of our proposed method comparing with conventional pruning methods and other once-for-all compression methods on ImageNet classification and COCO object detection. Further ablation experiments are performed to obtain the best configuration of the proposed framework. And OFARPruning can be migrated to pruning a single network which proves the wide applicability of the framework.

### 4.1. ImageNet Classification

In this section, we test ImageNet dataset based on MobileNetv1, MobileNetv2, and ResNet50. Among them, MobileNetv1 and MobileNetv2 represent lightweight network, while ResNet50 represents large network. Besides, MobileNetv1 is a typical non-residual network, while the rest are residual networks. Under the same range dynamic constraints of FLOPs, we first perform pruning on 1.5x width MobileNetv1, using the same training configuration as US-Net and Mutual-Net. The Accuracy-FLOPs curve is shown as Fig. 5(a). It shows that OFARPruning has higher accuracy than US-Net (73.1% vs. 71.8%) and MutualNet (73.1% vs. 72.4%) under 570 MFLOPs. Besides, under the dynamic constraints of [50, 570] MFLOPs, OFARPruning also achieves higher accuracy than MutualNet. Besides, we conduct experiments on MobileNetv2 and ResNet50 with residual structure to further show that the network structure obtained by OFARPruning is significantly better than reducing the network width directly, and the resolution considered in the searched network can promote inter-network information flow which boosts all the networks' performance. Similarly, the 1.5x width MobileNetv2 and ResNet are pruned under the same training configuration, and the comparison results are shown in Fig. 5(b) and Fig. 5(c). We can see that OFARPruning outperforms US-Net (74.1% vs.

71.2% on MobileNetv2) and MutualNet (74.1% vs. 72.7% on MobileNetv2) under the same FLOPs constrain. As shown in Table 1 and Table 2, while PFARPruning is highly efficient, its accuracy is even higher than the state-of-the-art conventional pruning methods which are once training for single network.

### 4.2. COCO Object Detection

To verify the robustness and universal applicability of the OFARPruning method, we migrate it to object detection. The experiment settings in this section are consistent with the MutualNet settings, and the VGG-16 [27] is pretrained by OFARPruning. Different from MutualNet and the above classification experiments, we only perform inplace distillation with teaching assistant on the classification branch for simplicity. In the training process, the input resolution is 1000× $\{600, 480, 360, 240\}$, and mAP@0.50:0.05:0.95 is used as the performance evaluation metric. The comparison results with US-Net and MutualNet are shown in the Table 3. It shows that OFARPruning outperforms the above two methods under the same resource constraints. We also achieve a compact model with higher mAP than the full model which validates that OFARPruning method effectively optimizes each sub-network in the framework.

### 4.3. Single Network Pruning

OFARPruning can also be implemented to a single network. According to Equation 1, increasing the maximum pruning rate can achieve the pruning of a single network. We conduct experiments on Cifar10 and Cifar100 based on Wider-ResNet. The resolution is selected from $\{32, 28, 24, 20\}$ and the pruning rate is set [$\rho_i$ 0.05, $\rho_i$ + 0.05] which 0.05 is only to distinguish when sampling. After searching the corresponding structure, we also train it for 200 epochs. The result is compared in Table 4.

Table 5. Ablation experiments based on MobileNetv2. TA means teaching assistant, $lr_s$ means learning rate in the searching stage, $\rho_n$ means the maximum pruning rate. We choose different FLOPs levels to compare the results. - means corresponding networks are not involved in optimization through once training.

| Configuration | FLOPs | Top-1 Accuracy |
|---|---|---|
| With TA, $lr_s$=0.5, $\rho_n$=0.8 | 504M / 253M / 179M / 107M | 75.5% / 74.0% / 72.6% / 69.2% |
| Without TA, $lr_s$=0.5, $\rho_n$=0.8 | 504M / 253M / 179M / 107M | 75.2% / 73.8% / 72.3% / 68.3% |
| With TA, $lr_s$=0.3, $\rho_n$=0.8 | 504M / 253M / 179M / 107M | 74.6% / 73.0% / 71.5% / 68.0% |
| With TA, $lr_s$=0.1, $\rho_n$=0.8 | 504M / 253M / 179M / 107M | 74.0% / 72.6% / 71.1% / 67.6% |
| With TA, $lr_s$=0.5, $\rho_n$=0.6 | 504M / 253M / 179M / 107M | 75.7% / 74.3% / - / - |
| With TA, $lr_s$=0.5, $\rho_n$=0.4 | 504M / 253M / 179M / 107M | 76.0% / - / - / - |

### 4.4. Ablation Experiment

In this section, we conduct ablation experiments for each component of OFARPruning to explore the main reasons that OFARPruning can improve the performance of the pruning network, and to determine the best configuration of OFARPruning.

**Inplace distillation with teaching assistant or not.** US-Net and Mutual-Net adopt normal knowledge distillation to jointly train the networks sampled by sandwich rule. But we find that when we increase the maximum pruning rate, that is, increase the executable range, the network performance with large pruning rate will be unsatisfactory. We believe that the information extraction ability gap between the teacher network and the student network is too large, causing the student network unable to effectively obtain the output information of the teacher network. In order to confirm this conjecture, we conduct a comparative experiment based on MobileNetv2 on whether to use distillation with teaching assistant. It can be seen from Table 5 that the distillation with teaching assistant has a greater impact on the network with large pruning rate, and can improve the accuracy of the teacher network to a certain extent.

**Learning rate in searching stage.** Though OFARPruning uses very low energy and time consumption to find a compact network structure that is close to the winner tickets, we find that the hyperparameters (especially learning rate) used in the searching stage have a great impact on the final result through experiments. With a fixed number of searching epochs and other related configurations, we fairly train the networks searched with different learning rates, and the results are shown in Table 5. It is observed that large learning rate contributes a lot to the emergence of high-quality network structures and final performances of such structures are also better.

**Maximum Pruning Rate.** It can be seen from Equation 1 that the upper bound and lower bound network will have an impact on OFARPruning, moreover the lower bound network has a greater impact on the performance of each subnetwork. For this problem, we conduct experiments on the maximum pruning rate (lower bound network), and the results are shown in Table 5. When the maximum pruning rate is reduced, the performance of each network is improved, but the dynamic range of the model will be correspondingly narrowed.

### 5. Conclusion and Discussion

The paper introduces a simple and efficient pruning framework which takes input resolution into consideration during network pruning. We prove that structures close to winner tickets can be searched with low time and energy consuming in this pruning framework. After searching, we can obtain a cohort of compact networks adaptive to different input resolution to meet dynamic resource constraints of various edge devices through once joint optimization. Through image classification and object detection experiments, we show that OFARPruning has a significant increase in accuracy compared to the current once-for-all compression methods, and is efficiency compared to conventional pruning methods. Besides, we explore the factors affecting the accuracy of the framework through ablation experiment and determine the best configuration of the framework. In the future, we will make extensions to this framework, such as transferring this method to other visual tasks, low-bit quantization to accelerate on edge devices, and utilizing neural architecture search in the framework to explore higher quality architectures.